\documentclass[letterpaper]{article}
\usepackage[preprint]{aaai2027}
\usepackage[hyphens]{url}
\usepackage{graphicx}
\usepackage{natbib}
\usepackage{caption}
\usepackage{algorithm}
\usepackage{algorithmic}
\usepackage{newfloat}
\usepackage{amsmath}
\usepackage{amssymb}
\DeclareMathSizes{10}{9}{7}{6.5}
\usepackage{listings}
\DeclareCaptionStyle{ruled}{labelfont=normalfont,labelsep=colon,strut=off}
\floatstyle{ruled}
\newfloat{listing}{tb}{lst}{}
\floatname{listing}{Listing}
\usepackage{booktabs}

\title{B-Spline Embedded Structure Learning for 3D Tooth Segmentation}
\author{
    Xianghan Wei\textsuperscript{\rm 1},
    Jianwen Lou\textsuperscript{\rm 1},
    Zhiguo Lu\textsuperscript{\rm 1},
    Hairong Jin\textsuperscript{\rm 2},
    Haihua Zhu\textsuperscript{\rm 3}
}
\affiliations{
    \textsuperscript{\rm 1}School of Software Technology, Zhejiang University\\
    \textsuperscript{\rm 2}School of Computer and Computing Science, Hangzhou City University\\
    \textsuperscript{\rm 3}The Affiliated Stomatology Hospital, Zhejiang University School of Medicine
}

\begin{document}

\maketitle

\begin{abstract}
Accurate 3D tooth segmentation forms the cornerstone of digital dentistry, yet it remains a formidable challenge due to the inherent intricacy of real-world dentitions, such as crowding, misaligned teeth and high morphological similarity between adjacent teeth. To resolve this, we present B-Spline Embedded Structure Learning, a novel framework that distills the inherent sequential arrangement of teeth into a continuous structural constraint to regularize representation space. Our approach parameterizes the global dental topology by fitting a parametric B-spline trajectory to tooth centers, assigning each point a continuous structural embedding that forces the shared backbone to capture global arch organization. To fully exploit these embedded priors, we introduce a Structure-Aware Dynamic Classifier (SADC) to substitute rigid static templates with adaptive, case-calibrated decision boundaries. SADC regularizes dynamic prototype pooling via a localized Gaussian proximity gate and contextually co-evolves them through an attention block modeling spatial relations and bilateral symmetries across teeth. Extensive evaluations on the 3DTeethSeg22 benchmark demonstrate that our method establishes a new state-of-the-art accuracy with exceptional structural robustness and efficiency in computational overhead, markedly enhancing the model's capacity to handle complex dental configurations.
\end{abstract}

\noindent\textbf{Code:} \url{https://github.com/blackbird2003/TeethSplineSeg}

\section{Introduction}
\label{sec:introduction}
Accurate and robust 3D tooth segmentation from high-resolution intraoral scans is fundamental to modern digital dentistry, including computer-aided orthodontics, prosthesis design, and surgical simulation. Advances in geometric deep learning have enabled point- and mesh-based networks - including general backbones~\cite{qiPointNetDeepHierarchical2017,wangDynamicGraphCNN2019,zhaoPointTransformer2021} and dental-specific architectures targeting multiscale contexts, local boundaries, or foundation representations~\cite{xu3DToothSegmentation2019,cuiTSegNetEfficientAccurate2021,lianDeepMultiScaleMesh2020,wuTwoStageMeshDeep2022,zhengTeethGNNSemantic3D2023,krenmayrDilatedToothSegNetToothSegmentation2024a,jinTSRNetDualStreamNetwork2025,jinLearningCenterBoundaryaware2025,limToothGroupNetwork2022,lu3DTeethSAMTamingSAM2} - to automate this task. Despite improving segmentation accuracy, existing frameworks still struggle with precise boundary delineation and reliable tooth category discrimination in complex real-world dentitions, including severe crowding, malpositioned teeth, and morphological ambiguity within functional groups (e.g., adjacent molars).

Incorporating anatomical priors, such as invariant dental order and stable dental arch shape, offers a promising yet largely unexplored solution. Most notably, DArch~\cite{qiuDArchDentalArch2022} pioneered explicit spatial layout modeling by representing the dental curve with parametric Bézier lines, demonstrating that structural priors facilitate tooth instance localization. However, its spatial prior primarily optimizes input-stage point sampling statistics rather than directly regularizing downstream neural representations. This leaves an opportunity to map continuous trajectories and rigid sequential topology into a unified dense embedding space, improving structural robustness in complex cases.

Another common bottleneck is the reliance on globally shared, static classification weights. Such rigid decision boundaries use fixed templates to separate teeth across subjects, limiting adaptation to individual distribution shifts and abnormal dentitions. Although the Context-Aware Classifier (CAC)~\cite{tianLearningContextAwareClassifier2023a} was recently introduced in general computer vision to generate sample-conditioned decision boundaries, this dynamic classification philosophy remains unexplored in 3D tooth segmentation. Moreover, generic CAC operates within unconstrained feature manifolds and may generate anatomically implausible boundaries for complex or corrupted dental layouts without explicit spatial regularization.

To address these gaps, we propose a novel B-spline embedded structure learning paradigm for 3D tooth segmentation. Specifically, we use a continuous B-spline trajectory to model the dental arch and parameterize its global topology as point-wise structural embeddings within the neural representation space. These dense structural embeddings are integrated into a Structure-Aware Dynamic Classifier (SADC), which regularizes dynamic prototype pooling and replaces rigid static templates with adaptive, case-calibrated decision boundaries, improving classification robustness across complex cases. In summary, the core contributions of this work are three-fold:
\begin{itemize}
\item We introduce a novel B-spline embedded structure learning paradigm that models the dental arch trajectory to extract dense, point-wise structural embeddings, unifying continuous geometric variations with rigid sequential topology.
\item We propose the Structure-Aware Dynamic Classifier (SADC), a pioneering dental-native classifier that replaces rigid, static templates with dynamic, case-calibrated decision boundaries regularized by structural location and relational context.
\item Extensive evaluations on the 3DTeethSeg22 benchmark demonstrate that our framework achieves new state-of-the-art accuracy and exceptional structural robustness across complex cases while significantly reducing training and inference overhead.
\end{itemize}

\subsection{Related Work}
\label{sec:related_work}
\subsubsection{Segmentation via Geometry Learning}
Early 3D tooth segmentation methods relied on handcrafted geometric cues such as surface curvature, morphological skeletons, and harmonic fields~\cite{yuanSingleToothModeling2010,wuToothSegmentationDental2014}. To address their limited generalization and need for manual tuning in complex cases, data-driven geometric deep learning has introduced advanced point- and mesh-based backbones, including PointNet++~\cite{qiPointNetDeepHierarchical2017}, DGCNN~\cite{wangDynamicGraphCNN2019}, and Point Transformer~\cite{zhaoPointTransformer2021}. Based on these architectures, Xu et al.~\cite{xu3DToothSegmentation2019} projected mesh facet geometries into 2D feature maps for hierarchical inference. TSegNet~\cite{cuiTSegNetEfficientAccurate2021} simplified instance tracking by predicting coarse tooth centroids, while MeshSegNet~\cite{lianDeepMultiScaleMesh2020} and iMeshSegNet~\cite{wuTwoStageMeshDeep2022} incorporated graph-constrained dental mesh representations. Similarly, TeethGNN~\cite{zhengTeethGNNSemantic3D2023} regressed center offsets through dual-space graphs to suppress fused segments. Recent methods further improve boundary precision and instance isolation through stronger feature learning and local geometry refinement. ToothGroupNet~\cite{limToothGroupNetwork2022} combined Point Transformer features with two-stage boundary-aware clustering, while 3DTeethSAM~\cite{lu3DTeethSAMTamingSAM2} and TSegAgent~\cite{zhuangTSegAgentZeroShotTooth2026b} exploited 2D vision foundation models through 3D-2D projections.Despite continued accuracy improvements, these methods still struggle with precise boundary delineation and reliable tooth category discrimination in complex dentitions, including crowding, tooth displacement, and high morphological similarity between adjacent teeth.

\subsubsection{Structural Prior Integration}
Structural prior regularization offers a promising solution to these challenges, yet remains underexplored in 3D tooth segmentation. Most notably, DArch~\cite{qiuDArchDentalArch2022} pioneered explicit spatial layout modeling using parametric Bézier lines to represent the dental curve. Its results demonstrate that macro-structural priors benefit tooth centroid clustering and overall prediction accuracy. However, its spatial prior is mainly used to optimize input-stage point sampling statistics rather than directly regularizing downstream point-wise feature learning. This leaves substantial room to integrate global structural properties into neural representations and build a more robust embedding manifold for diverse dental anomalies.

\subsubsection{Context-adaptive Classification}
A common limitation of both generic geometric and prior-regularized 3D tooth segmentation methods is their reliance on globally shared, static classification weights. Such rigid decision boundaries use fixed templates across subjects, limiting adaptation to individual distribution shifts and abnormal dentitions. Although the Context-Aware Classifier (CAC)~\cite{tianLearningContextAwareClassifier2023a} addresses this bottleneck in general computer vision through sample-conditioned weight generation, this dynamic paradigm remains unexplored in 3D tooth segmentation. Moreover, generic CAC optimizes decision boundaries in unconstrained feature spaces without geometric awareness, making it susceptible to producing topologically invalid classification planes on distorted dental scans lacking explicit spatial constraints.


\section{Methodology}
\label{sec:method}

\begin{figure*}[!t]
    \centering
    \includegraphics[width=1\linewidth]{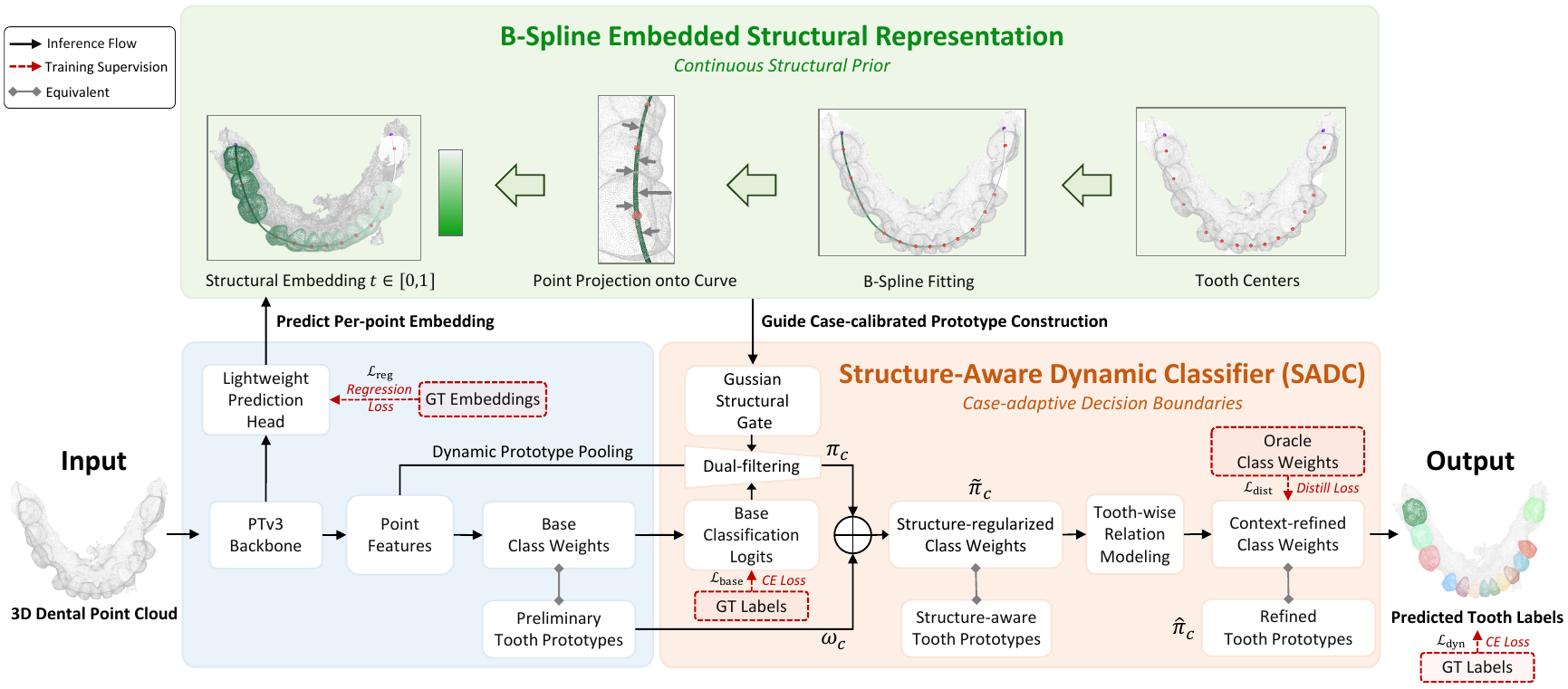}
    \caption{Overview of the proposed framework. A PTv3 backbone first extracts multi-scale geometric features from each dental arch. The B-spline embedded structural representation assigns each point a continuous structural coordinate $t$, while SADC uses these embeddings to construct and refine sample-specific tooth prototypes as dynamic classifier weights for final segmentation.}
    \label{fig:pipeline}
\end{figure*}

\subsection{Overview} 
\label{sec:overview} 
The overall framework is illustrated in Figure~\ref{fig:pipeline}. Given a 3D dental point cloud, we treat the topologically homogeneous maxillary and mandibular arches as independent samples and extract multi-scale features using a Point Transformer V3 (PTv3)~\cite{wuPointTransformerV32024c} backbone to capture local and long-range geometry. Our paradigm comprises two tightly coupled components. First, the \emph{B-Spline Embedded Structural Representation} models the stable sequential arrangement of teeth along the dental arch. We construct a tooth-center driven B-spline curve and project each point onto this continuous parametric trajectory to obtain a scalar coordinate $t$ as its point-wise structural embedding. A lightweight prediction head infers these coordinates, explicitly regularizing representation learning with the macroscopic 3D dentition layout. Second, the \emph{Structure-Aware Dynamic Classifier (SADC)} uses the predicted structural embeddings to construct sample-specific tooth prototypes. Instead of rigid static templates, SADC models spatial and structural relations among teeth and employs the refined prototypes as dynamic classifier weights. The final prediction thus integrates geometric feature learning with structural priors across complex cases.

\subsection{B-Spline Embedded Structural Representation}
\label{sec:structural_representation}
To encode the stable 3D dentition layout, we parameterize each dental arch as an ordered, continuous structural trajectory. Specifically, we fit a B-spline curve to tooth centers, project each 3D point onto the trajectory, and assign it a normalized, continuous coordinate. This parameter serves as a \emph{point-wise structural embedding} that regularizes downstream feature learning and subsequent Structure-Aware Dynamic Classifier (SADC) weight generation.

\subsubsection{Point-wise Structural Embedding}
For each single-arch sample with up to 16 valid tooth instances, the standard two-digit FDI notation is mapped to a normalized integer sequence $\mathcal{T}=\{1,\ldots,16\}$, ordered from the rightmost molar to the leftmost molar. We assign $c=0$ to the background/gingiva category, yielding the complete class set $\mathcal{C}=\{0\}\cup\mathcal{T}$. Given a labeled point cloud $\mathcal{P}=\{(p_n,y_n)\}_{n=1}^{N}$, where $p_n\in\mathbb{R}^{3}$ and $y_n\in\mathcal{C}$, the point subset and spatial centroid of each existing tooth category $c$ are defined as:
\begin{equation}
\begin{aligned}
    \mathcal{P}_c &= \{p_n \mid y_n=c\}, \\
    \mu_c &= \frac{1}{|\mathcal{P}_c|} \sum_{p_n\in\mathcal{P}_c} p_n, \quad c\in\mathcal{T}_{exist}
\label{eq:tooth_centroid}
\end{aligned}
\end{equation}
where $\mathcal{T}_{exist} \subseteq \mathcal{T}$ denotes the tooth categories present in the case. To connect this discrete sequence with continuous 3D geometry, each valid centroid $\mu_c$ is assigned a deterministic, normalized coordinate anchor $\tau_c = c/17$, where $c\in\mathcal{T}_{exist}$. This anchor provides an invariant reference encoding anatomical topology. Using the paired anchors $\{(\tau_c,\mu_c)\}$, we fit a cubic B-spline curve as a continuous 3D reference trajectory:
\begin{equation}
    C(t)=\sum_{k=0}^{K}N_{k,3}(t)P_k, \quad t\in[0,1]
\label{eq:bspline_curve}
\end{equation}
where $N_{k,3}(\cdot)$ denotes the cubic B-spline basis functions, and $P_k$ represents the control points obtained via algebraic interpolation. Missing teeth are excluded from 3D centroid interpolation, while their canonical locations remain reserved in the parameter space $[0,1]$. We also introduce virtual anchors at both arch boundaries to improve curve coverage and numerical stability near terminal molars. The fitted curve defines a continuous, subject-specific coordinate system across the dentition. For each input point $p_n$, its structural embedding $t_n\in[0,1]$ is obtained via orthogonal projection onto the B-spline trajectory:
\begin{equation}
    t_n = \arg\min_{t\in[0,1]} \left\|p_n-C(t)\right\|_2^2
\label{eq:structural_embedding}
\end{equation}
The resulting point-wise embedding preserves the macroscopic 3D spatial layout of the target dentition.

\subsubsection{Structural Embedding Prediction}
To incorporate the continuous anatomical prior into representation learning, we attach a lightweight auxiliary head to an intermediate PTv3 decoder stage. Given the feature vector $h_n$ extracted at point $p_n$, the predicted point-wise structural embedding is computed as $\hat{t}_n = \mathrm{sigmoid}(\mathrm{MLP}(h_n))$. This dense prediction head is supervised by the loss in Eq.~\eqref{eq:structural_loss}, encouraging the internal representation to encode the 3D spatial layout of the target dentition. The estimated coordinates $\hat{t}_n$ are then forwarded to SADC to guide prototype aggregation and inter-category relation modeling.

\subsection{Structure-Aware Dynamic Classifier}
\label{sec:sadc}
Conventional 3D tooth segmentation architectures use rigid classification layers with globally shared parameters, forcing fixed category descriptors to separate teeth across subjects and limiting adaptation to anomalous dental variations. SADC instead leverages the predicted point-wise structural embeddings $\hat{t}_n$ to construct a sample-conditioned classifier for each dentition. It aggregates structure-aware tooth prototypes from preliminary predictions via continuous proximity filters and refines them through sequence-wide relational modeling. These anatomically constrained prototypes then serve as dynamic weights in a cosine classification layer, replacing rigid static templates with adaptive, case-calibrated decision boundaries across complex cases.

\subsubsection{Structure-Aware Prototype Generation}
SADC extracts representative tooth prototypes ${\pi}_c \in \mathbb{R}^{64}$ ($c \in \mathcal{T}$) from unstructured point features to capture sample-specific geometric variations. To isolate reliable representations from noisy or ambiguous regions, we use a dual-filtering aggregation mechanism that jointly considers feature-space semantic confidence and B-spline-space structural proximity. Specifically, let $h_n \in \mathbb{R}^{64}$ be the PTv3 feature at point $p_n$. We project it into a segmentation feature space $f_n \in \mathbb{R}^{64}$ and compute the preliminary soft assignment $q_{n,c} = \operatorname{softmax}(w_c^{\top}f_n+b_c)$, where $w_c$ and $b_c$ are the globally shared class weight and bias of the base classifier. To remove non-surface noise and ambiguous boundaries, we construct a binary semantic mask $m_n = \mathbf{1}[\max_{c\in\mathcal{C}}q_{n,c}\geq\delta]$, retaining points whose confidence exceeds $\delta=0.75$. To reduce misassignments caused by local geometric similarities, we use the predicted point-wise structural embedding $\hat{t}_n$ to measure alignment with global arch topology. For each canonical tooth category $c \in \mathcal{T}$, we construct a continuous Gaussian structural gate - $G_{n,c} = \exp\left[ -\frac{1}{2} \left( \frac{\hat{t}_n-\tau_c}{\sigma_t} \right)^2 \right]$, where $\tau_c = c/17$ is the predefined canonical topological center of category $c$, and $\sigma_t=2/17$ controls the admissible bandwidth. The term $|\hat{t}_n-\tau_c|$ measures the parametric distance from the predicted embedding to the topological center of tooth $c$. The gate assigns higher affinity to points near the tooth center along the B-spline trajectory while suppressing anatomically distant regions. Combining the semantic mask and structural gate yields the point-wise aggregation weight $\alpha_{n,c} = m_n q_{n,c}G_{n,c}$. The dynamic feature prototype $\pi_c$ is then obtained via global weighted pooling:
\begin{equation}
    \pi_c = \frac{\sum_{n=1}^{N}\alpha_{n,c}f_n}{\sum_{n=1}^{N}\alpha_{n,c}+\epsilon}
\label{eq:initial_prototype}
\end{equation}
where $\epsilon$ is a small constant preventing division by zero. Finally, following ~\cite{tianLearningContextAwareClassifier2023a}, we fuse the dynamically pooled feature with its static counterpart through a lightweight fusion layer:
\begin{equation}
    \tilde{\pi}_c = \theta(\pi_c\mathbin{\Vert}w_c)
\label{eq:structure_refined_prototype}
\end{equation}
where $\mathbin{\Vert}$ denotes feature concatenation and $\theta$ is a linear projection network. This yields 16 structure-regularized prototypes containing both invariant global semantics and subject-specific anatomy for subsequent relation modeling.

\subsubsection{Tooth-Wise Relation Modeling}
The dynamic prototypes aggregated via Eq.~\eqref{eq:initial_prototype} incorporate structural constraints but are pooled independently, lacking explicit interactions among teeth. In practice, neighboring teeth form a continuous geometric chain, while bilaterally corresponding categories across the left and right quadrants exhibit strong morphological correlations. We therefore use a relation-aware self-attention network with geometric relational biases to propagate these dependencies across the 16 prototypes. We first estimate the sample-specific structural coordinate of each tooth. Using the normalized aggregation weights from Eq.~\eqref{eq:initial_prototype}, the refined parametric center for category $c$ is pooled as $\bar{t}_c = \sum_{n=1}^{N}\alpha_{n,c}\hat{t}_n / (\sum_{n=1}^{N}\alpha_{n,c}+\epsilon)$. If category $c$ is missing or lacks valid point-wise support, we set $\bar{t}_c=\tau_c$. For each category pair $(c,c')$, we construct a three-dimensional structural relation vector:
\begin{equation}
    r_{c,c'} = \left[ d_{c,c'},\, \left|d_{c,c'}-\frac{1}{17}\right|,\, \left|\bar{t}_c+\bar{t}_{c'}-1\right| \right]^\top
\label{eq:relation_feature}
\end{equation}
where $d_{c,c'} = |\bar{t}_c-\bar{t}_{c'}|$. The three components encode relative arch distance, deviation from the canonical adjacent interval, and deviation from bilateral symmetry, respectively. For each attention head $l \in \{1,\ldots,L\}$, the relation-augmented attention logit between categories $c$ and $c'$ is:
\begin{equation}
    a_{c,c'}^{l} = \frac{Q_l(\tilde{\pi}_c)^{\top}K_l(\tilde{\pi}_{c'})}{\sqrt{d_l}} + \varphi_l(r_{c,c'})
\label{eq:relation_attention}
\end{equation}
where $Q_l$ and $K_l$ are standard linear query and key projections, $d_l$ is the channel dimension per head, and $\varphi_l(\cdot)$ maps the geometric relation vector to a head-specific structural attention bias. We collect these logits into a sequence-wide attention map and pass the prototype token set $\tilde{\Pi} = [\tilde{\pi}_1; \ldots; \tilde{\pi}_{16}] \in \mathbb{R}^{16 \times 64}$ through a standard Transformer encoder layer. Guided by the structural relation bias, the ordered tokens exchange context to produce the context-refined prototype matrix:
\begin{equation}
    \hat{\Pi} = \operatorname{TransformerEncoder}\left(\tilde{\Pi}, \{a_{c,c'}^{l}\}\right)
\label{eq:context_refined_prototype}
\end{equation}
where the rows of the optimized matrix $\hat{\Pi} = [\hat{\pi}_1; \ldots; \hat{\pi}_{16}] \in \mathbb{R}^{16 \times 64}$ contain the refined tooth prototypes $\hat{\pi}_c$. This contextual message-passing captures local shape details, neighborhood distances, and global bilateral arch layouts.

\subsubsection{Dynamic Cosine Classification}
For final prediction, local point-wise features are mapped into the optimized prototype manifold for direct comparison in a unified embedding space. For each input point $p_n$, its projected feature is computed as $g_n=\psi(f_n)$, where $\psi(\cdot)$ represents a learnable feature projection layer. The context-refined prototype $\hat{\pi}_c$ ($c \in \mathcal{T}$) serves as the subject-specific dynamic classifier weight for its corresponding tooth category. The dynamic logit for each tooth category is computed using scaled cosine similarity:
\begin{equation}
    z_{n,c}^{\mathrm{dyn}} = s \cdot \frac{g_n^{\top}\hat{\pi}_c}{\|g_n\|_2\|\hat{\pi}_c\|_2}, \quad c\in\mathcal{T}
\label{eq:dynamic_classifier}
\end{equation}
where $s=15$ denotes an empirical scaling factor that calibrates the logit distribution. Because the gingiva/background category is not part of the ordered anatomical tooth sequence, SADC restricts contextual prototype refinement to the 16 valid tooth categories. To cover the complete scan, the dynamic tooth logits $z_{n,c}^{\mathrm{dyn}}$ are combined with the static background logit $z_{n,0}^{\mathrm{base}} = w_0^{\top}f_n+b_0$ computed by the base classifier (where $w_0$ and $b_0$ denote the static background weight and bias). The final classification outputs are:
\begin{equation}
    z_{n,c}^{\mathrm{final}} = 
    \begin{cases}
        z_{n,0}^{\mathrm{base}}, & c=0 \\[2pt]
        z_{n,c}^{\mathrm{dyn}}, & c\in\mathcal{T}
    \end{cases}
\label{eq:final_logits}
\end{equation}
The dense segmentation probability maps are obtained by applying a standard softmax function over $z_n^{\mathrm{final}}$. Thus, the background prediction remains anchored to a robust, globally shared baseline template, while each tooth decision boundary is dynamically conditioned on the subject-specific dentition geometry.

\subsection{Training Objectives}
\label{sec:training_objectives}
We jointly optimize the base feature backbone, point-wise structural prediction head, and SADC module. To stabilize prototype learning during early training, we introduce an oracle distillation branch constructed from ground-truth semantic assignments following~\cite{tianLearningContextAwareClassifier2023a}. The overall multi-task objective is:
\begin{equation}
    \mathcal{L} = \lambda_1\mathcal{L}_{\mathrm{base}} + \lambda_2\mathcal{L}_{\mathrm{dyn}} + \lambda_3\mathcal{L}_{\mathrm{reg}} + \lambda_4\mathcal{L}_{\mathrm{dist}}
\label{eq:total_loss}
\end{equation}
where $\lambda_1$, $\lambda_2$, $\lambda_3$, and $\lambda_4$ are hyper-parameters that balance the respective loss components.

\subsubsection{Segmentation Supervision}
The base and context-refined dynamic classifiers are independently supervised using standard cross-entropy losses to stabilize both static and dynamic decision boundaries:
\begin{equation}
\begin{aligned}
    \mathcal{L}_{\mathrm{base}} &= \frac{1}{N} \sum_{n=1}^{N} \operatorname{CE}(z_{n,\cdot}^{\mathrm{base}},y_n) \\ 
    \mathcal{L}_{\mathrm{dyn}} &= \frac{1}{N} \sum_{n=1}^{N} \operatorname{CE}(z_{n,\cdot}^{\mathrm{final}},y_n)
\label{eq:segmentation_losses}
\end{aligned}
\end{equation}
where $\operatorname{CE}(\cdot,\cdot)$ denotes point-wise cross-entropy, $y_n \in \mathcal{C}$ is the ground-truth semantic label of point $p_n$, and $z_{n,\cdot}^{\mathrm{base}}$, $z_{n,\cdot}^{\mathrm{final}} \in \mathbb{R}^{|\mathcal{C}|}$ are the full class logit vectors for point $p_n$.

\subsubsection{Structural Regularization}
Because the continuous point-wise structural embedding is defined only for valid anatomical teeth, we restrict supervision to the tooth point subset $\Omega_{\mathrm{tooth}} = \{n \mid y_n \in \mathcal{T}\}$. The regression objective is:
\begin{equation}
    \mathcal{L}_{\mathrm{reg}} = \frac{1}{|\Omega_{\mathrm{tooth}}|} \sum_{n\in\Omega_{\mathrm{tooth}}} \rho(\hat{t}_n-t_n)
\label{eq:structural_loss}
\end{equation}
where $t_n$ is the true B-spline projection parameter obtained from Eq.~\eqref{eq:structural_embedding}, and $\rho(\cdot)$ denotes a robust Smooth-$L_1$ penalty function. This objective encourages the shared backbone features to encode the dental arch topology.

\subsubsection{Oracle-Guided Distillation}
To bridge the training-inference discrepancy - where ground-truth and network-predicted labels are used for prototype construction, respectively - we adapt the entropy-aware knowledge distillation framework from~\cite{tianLearningContextAwareClassifier2023a}. During training, an oracle branch uses ground-truth tooth assignments to generate ideal reference prototypes $\hat{\pi}_c^{o}$. We then compute the entropy-weighted Kullback-Leibler (KL) divergence between the softened oracle distribution $\mathbf{P}_n^{o}$ and the refined prediction distribution $\mathbf{P}_n^{r}$ to enforce structural alignment:
\begin{equation}
\mathcal{L}_{\mathrm{dist}} = (T^2 / \sum_{n=1}^{N}\omega_{y_n}) \sum_{n=1}^{N} \omega_{y_n} D_{\mathrm{KL}}( \mathbf{P}_n^{o} \,\|\, \mathbf{P}_n^{r} )
\end{equation}
where $\omega_{y_n}$ denotes the class confidence weights evaluated from the oracle entropy following~\cite{tianLearningContextAwareClassifier2023a}.


\section{Experiments}

\subsection{Experimental Setup}
\subsubsection{Dataset}
We evaluate our framework on the widely adopted 3DTeethSeg22 benchmark~\cite{ben-hamadou3DTeethSeg223DTeeth2023}, comprising paired maxillary and mandibular high-resolution intraoral meshes from 900 subjects (1,800 meshes). All meshes have point-wise semantic annotations following the standard Federation Dentaire Internationale (FDI) notation. We use the official split of 1,200 meshes for training and 600 for testing, treating each arch as an independent sample.

\subsubsection{Metrics}
We use six standard metrics to assess local boundary alignment and global category identification: (1) Overall Accuracy (\textbf{OA}): Measures the ratio of correctly predicted mesh vertices, formulated as $\mathrm{OA} = \frac{1}{M}\sum_{n=1}^{M}\mathbf{1}(\hat{y}_n=y_n)$ for $M$ total vertices; (2) Teeth Mean Intersection over Union (\textbf{T-mIoU}): Measures overlap over existing tooth categories, defined as $\frac{1}{|\mathcal{T}_{exist}|}\sum_{i\in\mathcal{T}_{exist}} |P_i\cap G_i|/|P_i\cup G_i|$; (3) Teeth Dice Coefficient (\textbf{Dice}): Measures the harmonic mean of precision and recall over existing teeth via $\frac{1}{|\mathcal{T}_{exist}|}\sum_{i\in\mathcal{T}_{exist}} 2|P_i\cap G_i|/(|P_i|+|G_i|)$; (4) Boundary IoU (\textbf{B-IoU}): Gauges contour precision along clinically critical margins using a 10-neighborhood search to extract boundary sets $B_p$ and $B_g$, computed as $|B_p\cap B_g|/|B_p\cup B_g|$; (5) Tooth Identification Rate (\textbf{TIR}): Quantifies the ratio of correctly classified teeth whose centroid distance falls within half the target tooth's spatial diameter; (6) \textbf{TIR$_{=1}$}: Signifies the percentage of samples with error-free anatomical identification across all 16 canonical categories.

\subsection{Implementation Details}
\subsubsection{Model and Optimization Setup}
Our backbone follows the standard PTv3 configuration from its original release. The structural embedding prediction head is attached after the second PTv3 decoder stage, while SADC uses a 4-head relation-aware self-attention block to model the 16 tooth categories. All models are evaluated on a single NVIDIA GeForce RTX 4090 GPU. We train for 100 epochs with a batch size of 8 using Adam with $\beta = (0.9, 0.999)$ and a weight decay of $10^{-5}$. The learning rate starts at $10^{-3}$, with a linear warm-up over the first 5 epochs followed by cosine annealing to $10^{-5}$. The loss weights are set to $\lambda_1=1.0$, $\lambda_2=1.0$, $\lambda_3=100.0$, and $\lambda_4=0.5$.

\begin{table*}[!t]
\centering
\begin{tabular}{lcccccc}
\toprule
Method & OA$\uparrow$\ & T-mIoU$\uparrow$\ & Dice$\uparrow$\ & B-IoU$\uparrow$\ & TIR$\uparrow$\ & TIR$_{=1}$$\uparrow$\ \\
\midrule
TSegAgent~\cite{zhuangTSegAgentZeroShotTooth2026b} & 62.16 & 31.45 & 32.52 & 64.38 & 34.38 & 14.67 \\
TeethGNN~\cite{zhengTeethGNNSemantic3D2023} & 90.96 & 83.89 & 88.46 & 48.49 & 93.78 & 73.33 \\
ISBNet~\cite{ngoISBNet2023} & 91.53 & 81.01 & 87.65 & 39.61 & 96.49 & 89.00 \\
CBANet~\cite{jinLearningCenterBoundaryaware2025} & 92.72 & 86.63 & 91.03 & 50.73 & 96.44 & 86.67 \\
TSRNet~\cite{jinTSRNetDualStreamNetwork2025} & 92.87 & 86.56 & 90.91 & 51.11 & 96.09 & 87.17 \\
DilatedToothSegNet~\cite{krenmayrDilatedToothSegNetToothSegmentation2024a} & 93.40 & 86.55 & 91.26 & 51.57 & 96.80 & 89.33 \\
Point Transformer V3~\cite{wuPointTransformerV32024c} & 94.45 & 89.02 & 92.48 & 63.67 & 96.53 & 89.17 \\
ToothGroupNet~\cite{limToothGroupNetwork2022} & \underline{95.19} & 90.16 & 92.88 & \underline{69.30} & 96.83 & 88.83 \\
3DTeethSAM~\cite{lu3DTeethSAMTamingSAM2} & \underline{95.19} & \underline{91.44} & \underline{94.01} & 69.06 & \underline{96.99} & \underline{90.83} \\
Ours & \textbf{96.01} & \textbf{92.62} & \textbf{94.90} & \textbf{71.36} & \textbf{97.50} & \textbf{92.67} \\
\bottomrule
\end{tabular}
\caption{Quantitative results on the test set (\%).}
\label{tab:comparison_results}
\end{table*}

\begin{table}[h]
\centering
\resizebox{\columnwidth}{!}{
\begin{tabular}{cccl}
\toprule
Rank & Count & Binary Vector & Missing Tooth \\
\midrule
1 & 334 & \texttt{0111111111111110} & 1, 16 \\
2 & 104 & \texttt{0011111111111100} & 1, 2, 15, 16 \\
3 & 29 & \texttt{0011111111111110} & 1, 2, 16 \\
4 & 22 & \texttt{1111111111111111} & None \\
5 & 21 & \texttt{0111111111111100} & 1, 15, 16 \\
Others & $\leq 8$ each, 90 in total & -  & - \\
\bottomrule
\end{tabular}
}
\caption{Tooth-presence configurations in the test set.}
\label{tab:top_presence_patterns}
\end{table}

\subsubsection{Data Preprocessing and Postprocessing}
Following established dental segmentation protocols~\cite{jinTSRNetDualStreamNetwork2025}, high-resolution meshes are downsampled before network input to meet the memory limits of 3D deep networks. For each dental arch mesh, we sample $N=32{,}000$ points using Farthest Point Sampling (FPS). The coordinates are zero-centered, normalized by the maximum spatial diameter. For fair comparison, predictions from all baseline methods are upsampled to the dense full-resolution mesh using the same nearest-neighbor interpolation before evaluation. To remove stray fragments and refine boundaries, we apply a graph-cut smoothing pipeline adapted from 3DTeethSAM~\cite{lu3DTeethSAMTamingSAM2}. A sparse vertex adjacency graph is constructed from the original mesh faces to reassign disconnected component islands to the background class, unless a secondary component exceeds half the primary component's scale and lies within half its spatial distance. Finally, a mesh graph-cut module initialized with the cleaned labels performs tooth-wise fuzzy clustering to regularize the final boundaries without altering the network weights.

\subsection{Comparison with State-of-the-Art Methods}
\label{sec}
We compare our framework with representative state-of-the-art (SOTA) methods on the 3DTeethSeg22 benchmark, including graph neural networks, Transformers, and foundation-model-adapted paradigms. 

\subsubsection{Segmentation Accuracy Evaluation}
As shown in Table~\ref{tab:comparison_results}, our method establishes a new SOTA, ranking first across all metrics. Despite its compact native 3D architecture, it consistently surpasses the foundation-model-based 3DTeethSAM by up to $2.30\%$ and the strongest native 3D baseline, ToothGroupNet, by up to $3.84\%$. Relative to the plain PTv3 backbone, our structural modeling yields particularly substantial gains in boundary delineation and perfect tooth identification ($+7.69\%$ on B-IoU and $+3.50\%$ on TIR$_{=1}$), confirming that the improvements arise from B-spline structural regularization and SADC rather than backbone scaling.

\begin{table*}[!t]
\centering
\resizebox{\linewidth}{!}{
\begin{tabular}{lccccccccc}
\toprule
Method & \multicolumn{3}{c}{Top-1 Pattern (334)} & \multicolumn{3}{c}{Rank 2--5 Patterns (176)} & \multicolumn{3}{c}{Rare Patterns (90)} \\
\cmidrule{2-10}
 & T-mIoU$\uparrow$\ & TIR$\uparrow$\ & TIR$_{=1}$$\uparrow$\ & T-mIoU$\uparrow$\ & TIR$\uparrow$\ & TIR$_{=1}$$\uparrow$\ & T-mIoU$\uparrow$\ & TIR$\uparrow$\ & TIR$_{=1}$$\uparrow$\ \\
\midrule
TeethGNN~\cite{zhengTeethGNNSemantic3D2023} & 88.73 & 98.99 & 94.01 & 80.48 & 90.61 & 53.41 & 72.60 & 80.64 & 35.56 \\
ISBNet~\cite{ngoISBNet2023} & 83.84 & 99.30 & 97.60 & 82.33 & \underline{98.95} & \underline{93.18} & 67.95 & 81.23 & 48.89 \\
CBANet~\cite{jinLearningCenterBoundaryaware2025} & 89.55 & 99.57 & 97.31 & 88.38 & 98.33 & 91.48 & 72.39 & 81.13 & 37.78 \\
TSRNet~\cite{jinTSRNetDualStreamNetwork2025} & 89.81 & 99.45 & 97.31 & 88.19 & 98.37 & 90.91 & 71.30 & 79.13 & 42.22 \\
DilatedToothSegNet~\cite{krenmayrDilatedToothSegNetToothSegmentation2024a} & 89.45 & 99.68 & 98.50 & 87.47 & 98.12 & 89.20 & 74.01 & 83.55 & 55.56 \\
Point Transformer V3~\cite{wuPointTransformerV32024c} & 92.20 & 99.52 & 98.50 & 89.48 & 97.45 & 89.20 & 76.30 & 83.64 & 54.44 \\
ToothGroupNet~\cite{limToothGroupNetwork2022} & 92.64 & 99.42 & 97.31 & 91.74 & 98.85 & 90.34 & 77.87 & 83.26 & 54.44 \\
3DTeethSAM~\cite{lu3DTeethSAMTamingSAM2} & \underline{94.57} & \textbf{99.87} & \textbf{99.70} & \underline{92.27} & 98.01 & 90.91 & \underline{78.22} & \underline{84.32} & \underline{57.78} \\
Ours & \textbf{94.86} & \underline{99.73} & \underline{99.40} & \textbf{94.20} & \textbf{99.32} & \textbf{96.59} & \textbf{81.24} & \textbf{85.66} & \textbf{60.00} \\
\bottomrule
\end{tabular}
}
\caption{Comparison across frequency-stratified tooth-presence pattern groups (\%). Our pronounced advantage on rare patterns demonstrates superior robustness and generalization.}
\label{tab:pattern_group_results}
\end{table*}

\begin{figure*}[!t]
\centering
\includegraphics[width=\textwidth]{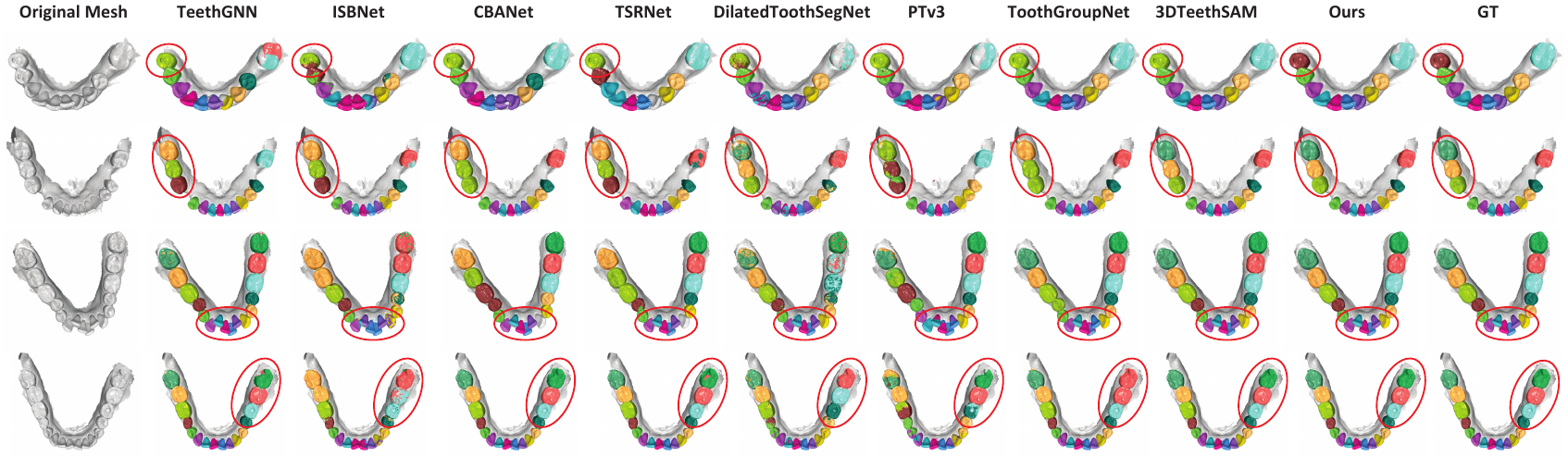}
\caption{Visual comparison on challenging test samples. Our method demonstrates superior boundary delineation and tooth identification in cases with missing teeth, severe crowding, and malalignment.}
\label{fig:visual_comparison}
\end{figure*}

To assess generalization beyond dominant tooth-presence patterns, we group the test cohort by missing-tooth configurations (Table~\ref{tab:top_presence_patterns}). As shown in Table~\ref{tab:pattern_group_results}, our framework ranks first on the dominant Top-1 pattern, with its advantage increasing as configurations become rarer. For the Rank 2--5 cohorts, it improves T-mIoU from 3DTeethSAM's $92.27\%$ to $94.20\%$ and achieves $96.59\%$ TIR$_{=1}$ ($+3.41\%$ over the runner-up). On "Rare Patterns" ($\leq 8$ cases each), it further leads the strongest baseline by $3.02\%$ in T-mIoU, $1.34\%$ in TIR, and $2.22\%$ in TIR$_{=1}$. The widening gains as configurations become increasingly abnormal demonstrate the superior structural robustness of our dynamic classifier.

The quantitative gains are corroborated by the visual results in Figure~\ref{fig:visual_comparison}. While previous methods suffer from boundary bleeding and category confusion under anatomical anomalies, our framework produces complete, distinct, and position-consistent tooth segments, maintaining strong topological consistency under severe crowding, pathological displacement, and missing teeth.

\subsubsection{Computational Efficiency Analysis}
Built on PTv3, our framework introduces structural reasoning through lightweight MLPs, point-wise prototype pooling, and relation modeling over only 16 tokens, avoiding foundation models and multi-stage grouping networks. As shown in Table~\ref{tab:efficiency_results}, on a single NVIDIA RTX 4090 GPU, training takes 2\,h 15\,m and the full inference pipeline takes 0.84\,s per scan. Compared with 3DTeethSAM and ToothGroupNet, our framework trains 9.2$\times$ and 9.8$\times$ faster and performs inference 4.1$\times$ and 3.5$\times$ faster, respectively. These results demonstrate that our framework substantially reduces training and deployment costs while achieving superior segmentation precision.

\begin{table}[!t]
\centering
\resizebox{\columnwidth}{!}{
\begin{tabular}{lccc}
\toprule
Method & Training Time & Total Inference Time & Avg. / Scan \\
\midrule
3DTeethSAM & 20 h 36 m & 34 m 29 s & 3.45 s \\
ToothGroupNet & 22 h 9 m & 29 m 6 s & 2.91 s \\
\textbf{Ours} & \textbf{2 h 15 m} & \textbf{8 m 21s} & \textbf{0.84 s} \\
\bottomrule
\end{tabular}
}
\caption{Computational efficiency comparison.}
\label{tab:efficiency_results}
\end{table}

\begin{table}[!t]
\centering
\resizebox{\columnwidth}{!}{
\begin{tabular}{lcccccc}
\toprule
Setting & OA$\uparrow$\ & T-mIoU$\uparrow$\ & Dice$\uparrow$\ & B-IoU$\uparrow$\ & TIR$\uparrow$\ & TIR$_{=1}$$\uparrow$\ \\
\midrule
Full model & \textbf{96.01} & \textbf{92.62} & \textbf{94.90} & \textbf{71.36} & \textbf{97.50} & \textbf{92.67} \\
w/o BSS & 95.82 & 92.28 & 94.59 & 71.16 & 97.18 & 91.83 \\
w/o SADC & 95.73 & 92.10 & 94.46 & 70.81 & 97.17 & 91.33 \\
w/o BSS \& SADC  & 94.76 & 89.59 & 92.49 & 67.85 & 96.77 & 88.50 \\
w/o SADC's GSS & 95.83 & 92.30 & 94.57 & 71.35 & 97.15 & 91.61 \\
w/o SADC's RAL & 95.89 & 92.37 & 94.68 & 71.22 & 97.36 & 92.00 \\
\bottomrule
\end{tabular}
}
\caption{Ablation study of core components (\%). ``BSS'' denotes continuous B-spline structural supervision; ``GSS'' and ``RAL'' denote SADC's Gaussian structural gate and relation attention layer, respectively.}
\label{tab:ablation_results}
\end{table}

\subsection{Ablation Studies}
\label{sec:ablation_studies}
We ablate B-spline structural supervision, SADC, and its Gaussian structural gate and relation attention layer.

\subsubsection{Impact of Structural Supervision and SADC}
As shown in Table~\ref{tab:ablation_results}, both B-spline structural supervision and SADC are critical for modeling global dental constraints. Removing SADC reduces T-mIoU from $92.62\%$ to $92.10\%$ and B-IoU from $71.36\%$ to $70.81\%$, validating case-adaptive prototypes over fixed classification templates. Removing continuous B-spline structural supervision consistently degrades T-mIoU, Dice, B-IoU, and TIR, showing that centroid-anchored arch position regression promotes global topology encoding. Removing both causes the largest decline, reducing T-mIoU and TIR$_{=1}$ to $89.59\%$ and $88.50\%$, respectively. These results confirm their complementarity: structural supervision learns regularized representation, while SADC constructs case-calibrated prototypes within this structured feature space.

\subsubsection{Synergy of Dynamic Classifier Refinements}
As shown in Table~\ref{tab:ablation_results}, either refinement alone is suboptimal. The Gaussian structural gate enforces anatomically plausible prototype regions but may restrict adaptation to global morphology, whereas relation attention captures sequential context but may over-smooth adjacent or bilaterally symmetric prototypes without spatial constraints. Combining them achieves the best performance, demonstrating their complementarity in spatial regularization and contextual calibration.

\section{Limitations and Future Work}
\label{sec:limitations}
Our framework is constrained by the limited throughput of current 3D networks. High-resolution dental meshes must be downsampled, with predictions projected back to the original resolution and refined through post-processing, preventing fully end-to-end segmentation and adding computational overhead. Future work will explore high-throughput 3D architectures that directly process dense meshes while preserving fine-grained boundaries.

It is also worth noting that emerging foundation-model-based paradigms, such as TSegAgent~\cite{zhuangTSegAgentZeroShotTooth2026b}, offer an alternative route to native 3D segmentation by combining multi-view rendering, zero-shot instance segmentation with SAM3~\cite{carion2026sam3segmentconcepts}, and vision-language reasoning~\cite{seed2026seed18modelcardgeneralized}. However, its open-source implementation achieves only $62.16\%$ OA and $31.45\%$ T-mIoU on 3DTeethSeg22, with an average inference time of 372.5\,s per sample. Future work will explore more efficient integration of pretrained vision-language models through feature distillation, or reduced-round reasoning, while preserving the accuracy and efficiency of native 3D representations.

\section{Conclusion}
\label{sec:conclusion}
We present a novel B-spline embedded structure learning framework for addressing boundary bleeding and category confusion in 3D tooth segmentation. By parameterizing the dental arch as a continuous 3D trajectory, the framework derives point-wise structural embeddings that connect anatomical topology with semantic prediction. The proposed Structure-Aware Dynamic Classifier (SADC) uses these embeddings to construct and relationally refine subject-adaptive tooth prototypes, replacing static templates with dynamic, case-calibrated decision boundaries. Extensive evaluations on 3DTeethSeg22 demonstrate new state-of-the-art accuracy, strong robustness to irregular dental configurations, and substantial efficiency gains.

\bibliography{ref}


\end{document}